\documentclass{article}
\usepackage{spconf,amsmath,graphicx}
\usepackage{hyperref}

\usepackage{booktabs}
\usepackage{xcolor,colortbl}
\usepackage{tikz}

\title{Semi-Supervised Spoken Language Understanding\\
via Self-Supervised Speech and Language Model Pretraining}

\name{Cheng-I Lai$^{\star}$, Yung-Sung Chuang$^{\dagger}$, Hung-Yi Lee$^{\dagger}$, Shang-Wen Li\thanks{The fourth author contributed to the work before joining Amazon.}$^{\ddagger}$, James Glass$^{\star}$}

\address{
        $^{\star}$ MIT Computer Science and Artificial Intelligence Laboratory \\
	    $^{\dagger}$ College of Electrical Engineering and Computer Science, National Taiwan University %\\
	    $^{\ddagger}$ Amazon AI
}

\usepackage{amssymb}% http://ctan.org/pkg/amssymb
\usepackage{pifont}% http://ctan.org/pkg/pifont
\usepackage{amsmath,bm,multicol,multirow}
\usepackage{tablefootnote}
\usepackage{longtable}

\begin{document}
\ninept
\maketitle

\vspace{-2mm}
\begin{abstract}
\vspace{-2mm}
Much recent work on Spoken Language Understanding (SLU) is limited in at least one of three ways: models were trained on oracle text input and neglected ASR errors, models were trained to predict only intents without the slot values, or models were trained on a large amount of in-house data.
In this paper, we propose a clean and general framework to learn semantics directly from speech with semi-supervision from transcribed or untranscribed speech to address these issues.
Our framework is built upon pretrained end-to-end (E2E) ASR and self-supervised language models, such as BERT, and fine-tuned on a limited amount of target SLU data.
We study two semi-supervised settings for the ASR component: supervised pretraining on transcribed speech, and unsupervised pretraining by replacing the ASR encoder with self-supervised speech representations, such as wav2vec. 
In parallel, we identify two essential criteria for evaluating SLU models: environmental noise-robustness and E2E semantics evaluation.  
Experiments on ATIS show that our SLU framework with speech as input can perform on par with those using oracle text as input in semantics understanding, even though environmental noise is present and a limited amount of labeled semantics data is available for training. 

\end{abstract}


\vspace{-2mm}
\section{Introduction}
\label{sec:intro}
\vspace{-2.5mm}
Spoken Language Understanding (SLU)\footnote{SLU typically consists of Automatic Speech Recognition (ASR) and Natural Language Understanding (NLU). ASR maps audio to text, and NLU maps text to semantics. Here, we are interested in learning a mapping directly from raw audio to semantics.} is at the front-end of many modern intelligent home devices, virtual assistants, and socialbots \cite{yu2019gunrock,coucke2018snips}: given a spoken command, an SLU engine should extract relevant semantics\footnote{Semantic acquisition is commonly framed as Intent Classification (IC) and Slot Labeling/Filling (SL), see \cite{yu2019gunrock,coucke2018snips,tur2010left}.} from spoken commands for the appropriate downstream tasks.
Since SLU tasks such as the Airline Travel Information System (ATIS) \cite{hemphill1990atis}, the field has progressed from knowledge-based \cite{seneff1992tina} to data-driven approaches, notably those based on neural networks. 
In the seminal paper on ATIS by Tur et al. \cite{tur2010left}, incorporating linguistically motivated features for NLU and improving ASR robustness were underscored as the research emphasis for the coming years. 
Now, a decade later, we should ask ourselves again, how much has the field progressed, and what is left to be done? 

    \begin{figure}[!htbp]
    \centering
    \includegraphics[width=0.9\linewidth]{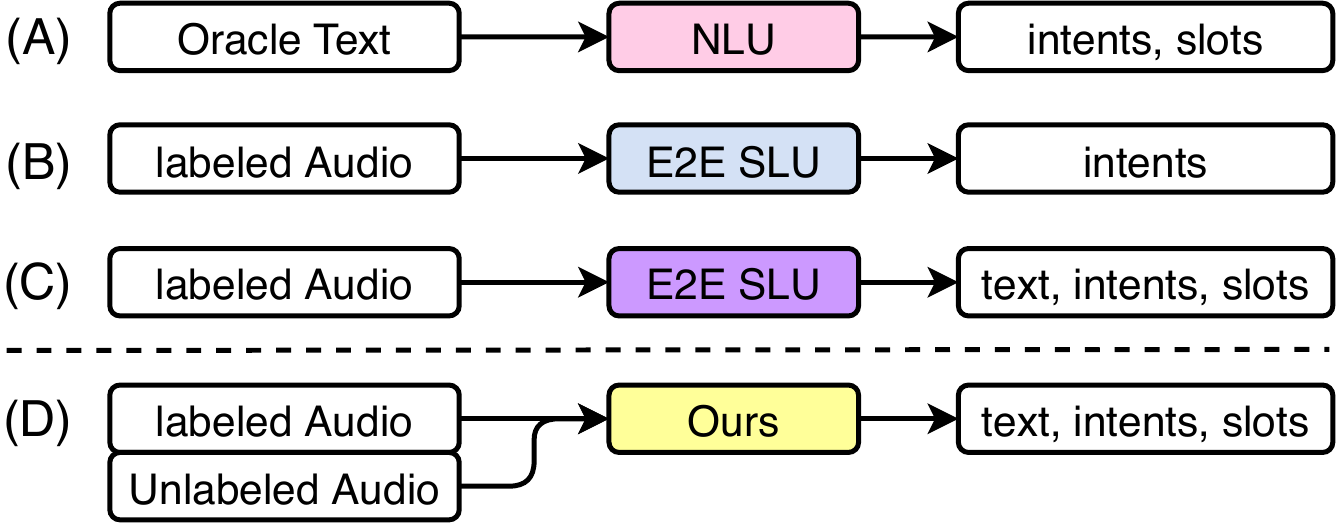}
    \vspace{-2.5mm}
    \caption{Comparison of input/output pairs of our proposed framework with past work, which are categorized as one of: 
    \textbf{(A)} NLU, which assumes oracle text as input instead of speech,
    \textbf{(B)} predicting \textit{intent only} from speech, ignoring their slot values, and
    \textbf{(C)} predicting text, intent, and slots from speech.
    \textbf{(D)} Our work predicts text, intent, and slots from speech while taking advantage of \textit{unlabeled} data.}
    \label{fig:teaser}
    \vspace{-6mm}
    \end{figure}
    
Self-supervised language models (LMs), such as BERT \cite{devlin2018bert}, and end-to-end SLU \cite{haghani2018audio,serdyuk2018towards,lugosch2019speech} appear to have addressed the problems posed in \cite{tur2010left}.
As shown in Figure~\ref{fig:teaser}, we can examine past SLU work from the angle of how they constructed the input/output pairs.
% We generally refer to the two approaches as \textbf{discriminative training} in the context of SLU, given that the core methodology is the same and differ mainly by their input/output pairs, see Figure~\ref{fig:teaser}.
In \cite{chen2019bert}, Intent Classification (IC) and Slot Labeling (SL) are jointly predicted on top of BERT, discarding the need of a Conditional Random Fields (CRF) \cite{zhou2015end}. 
However, these NLU works \cite{chen2019bert,zhu2017encoder,huang2020learning} usually ignore ASR or require an off-the-shelf ASR during testing. 
A line of E2E SLU work does take speech as input, yet it frames slots as intents and therefore their SLU models are really designed for IC only \cite{serdyuk2018towards,lugosch2019speech,wang2020large,cho2020speech,radfar2020end}. 
Another line of E2E SLU work jointly predicts text and IC/SL from speech, yet it either requires large amounts of in-house data, or restricts the pretraining scheme to ASR subword prediction \cite{haghani2018audio,rao2020speech,tomashenko2019recent,ghannay2018end}. 
In contrast, we would desire a framework that predicts text, intents, and slots from speech, while learning with limited semantics labels by pretraining on unlabeled data.

    \begin{figure*}[!htbp]
    \centering
    \includegraphics[width=1.0\linewidth]{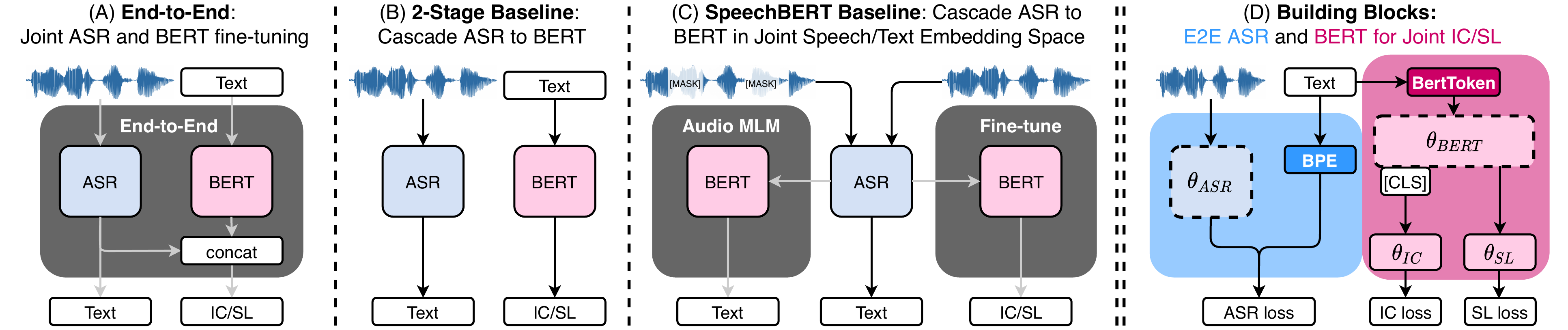}
    \vspace{-6mm}
    \caption{Our proposed semi-supervised learning framework with ASR and BERT for joint intent classification (IC) and slot labeling (SL) directly from speech. 
    \textbf{(A)} shows the end-to-end approach, in which E2E ASR and BERT are trained jointly by predicting text and IC/SL. 
    \textbf{(B)} shows the 2-Stage baseline, where text and IC/SL are obtained successively.
    \textbf{(C)} shows the SpeechBERT baseline, where BERT is adapted to take audio as input by first pretraining with Audio MLM loss and then fine-tuning for IC/SL. 
    A separate pretrained ASR is still needed for (B) and (C).
    \textbf{(D)} shows the ASR ($\theta_{ASR}$) and NLU ($\{\theta_{BERT}, \theta_{IC}, \theta_{SL}\}$) building blocks used in (A)-(C). 
    Note that $\theta_{ASR}$ and $\theta_{BERT}$ have different subword tokenizations: SentencePiece (BPE) \cite{kudo2018sentencepiece} and BertToken.
    Dotted shapes are pretrained.
    Figure best viewed in colors.}
    \label{fig:overview}
    \vspace{-5mm}
    \end{figure*}

\textbf{The case for semi-supervised SLU.}
Neural networks benefit from large quantities of labeled training data, and one can train end-to-end SLU models with them \cite{coucke2018snips,haghani2018audio,serdyuk2018towards,rao2020speech}. 
However, curating labeled IC/SL data is expensive, and often only a limited amount of labels are available. 
Semi-supervised learning could be a useful scenario for training SLU models for various domains whereby model components are pretrained on large amounts of unlabeled data and then fine-tuned with target semantic labels. 
While \cite{lugosch2019speech,wang2020large,rao2020speech,tomashenko2019recent} have explored this pretraining then fine-tuning scheme, they did not take advantage of the generalization capacity of contextualized LMs, such as BERT, for learning semantics from speech.
Notably, self-supervised speech representation learning \cite{chung2019unsupervised,oord2018representation,baevski2020wav2vec,schneider2019wav2vec,liu2020mockingjay} provides a clean and general learning mechanism for downstream speech tasks, yet the semantic transferrability of these representations are unclear.
Our focus is on designing a better learning framework distinctly for \textit{semantic understanding} under limited semantic labels, on top of ASR and BERT. 
We investigated two learning settings for the ASR component: (1) pretraining on transcribed speech with ASR subword prediction, and (2) pretraining on untranscribed speech data with contrastive losses \cite{baevski2020wav2vec,schneider2019wav2vec}. 

\vspace{1mm}\noindent
The key contributions of this paper are summarized as follows:
\vspace{-1mm}
\begin{itemize}
  \item We introduce a semi-supervised SLU framework for learning semantics from speech to alleviate: 
  (1) the need for a large amount of in-house, homogenous data \cite{coucke2018snips,haghani2018audio,serdyuk2018towards,rao2020speech}, 
  (2) the limitation of only intent classification \cite{serdyuk2018towards,lugosch2019speech,huang2020learning} by predicting text, slots and intents, and 
  (3) any additional manipulation on labels or loss, such as label projection \cite{caostyle}, output serialization \cite{haghani2018audio,tomashenko2019recent,ghannay2018end}, ASR n-best hypothesis, or ASR-robust training losses \cite{huang2020learning,lee2019mitigating}. 
  Figure~\ref{fig:overview} illustrates our approach.
    
  \item We investigate two learning settings for our framework: supervised pretraining and unsupervised pretraining (Figure~\ref{fig:wav2vec_learning}), and evaluated our framework with a new metric, the slot edit $F_{1}$ score, for end-to-end semantic evaluation. 
  Our framework improves upon previous work in Word Error Rate (WER) and IC/SL on ATIS, and even rivaled its NLU counterpart with oracle text input \cite{chen2019bert}.
  In addition, it is trained with noise augmentation such that it is robust to real environmental noises.
\end{itemize}

\vspace{-4mm}
\section{Proposed Learning Framework}
\label{sec:learning}
\vspace{-2mm}
We now formulate the mapping from speech to text, intents, and slots.
Consider a target SLU dataset $\mathcal{D} = \{\bm{A},\bm{W},\bm{S}, \bm{I}\}_{i=1}^{M}$, consisting of $M$ i.i.d. sequences, where $\bm{A},\bm{W},\bm{S}$ are the audio, word and slots sequences, and $\bm{I}$ is their corresponding intent label. 
Note that $\bm{W}$ and $\bm{S}$ are of the same length, and $\bm{I}$ is a one hot vector.  
We are interested in finding the model $\theta_{SLU}^*$ with loss, 
    \vspace{-1mm}
    \begin{equation}
    \mathcal{L}_{SLU}(\theta_{SLU}) =
    \mathop{\mathbb{E}_{(\bm{A},\bm{W},\bm{S},\bm{I})\sim\mathcal{D}}}
    \Big[ 
    \ln P(\bm{W},\bm{S},\bm{I} \mid \bm{A};\theta) 
    \Big]
    \end{equation}
    
We proceed to describe an end-to-end implementation of $\theta_{SLU}$\footnote{We abuse some notations by representing models by their model parameters, e.g. $\theta_{ASR}$ for the ASR model and $\theta_{BERT}$ for BERT.}.

\vspace{-3mm}
\subsection{End-to-End: Joint E2E ASR and BERT Fine-Tuning.}
\label{subsec:end_to_end}
\vspace{-1.5mm}
    As illustrated in Figure~\ref{fig:overview}, $\theta_{SLU}$ consists of a pretrained E2E ASR $\theta_{ASR}$ and a pretrained deep contextualized LM $\theta_{NLU}$, such as BERT, and is fine-tuned jointly for $\bm{W}$, $\bm{S}$ and $\bm{I}$ on $\mathcal{D}$.
    The choice of E2E ASR over hybrid ASR here is because the errors from $\bm{S}$ and $\bm{I}$ can be back-propagated through $\bm{A}$; following \cite{chen2019bert}, we have $\bm{S}$ predicted via an additional CRF/linear layer on top of BERT, and $\bm{I}$ is predicted on top of the BERT output of the [CLS] token.
    The additional model parameters for predicting SL and IC are $\theta_{SL}$ and $\theta_{IC}$, respectively, and we have $\theta_{NLU} = \{\theta_{BERT}, \theta_{IC}, \theta_{SL}\}$.
    During end-to-end fine-tuning, outputs from $\theta_{ASR}$ and $\theta_{BERT}$ are concatenated to predict $\bm{S}$ and $\bm{I}$ with loss $\mathcal{L}_{NLU}$, while $\bm{W}$ is predicted with loss $\mathcal{L}_{ASR}$.
    The main benefit this formulation brings is that now $\bm{S}$ and $\bm{I}$ do not solely depend on an ASR top-1 hypothesis $\bm{W}^*$ during training, and the end-to-end objective is thus,
        \begin{equation}
        \vspace{-1.5mm}
        \mathcal{L}_{SLU}(\theta_{SLU}) = \mathcal{L}_{ASR}(\theta_{SLU}) + \mathcal{L}_{NLU}(\theta_{SLU}).
        \vspace{1.5mm}
        \end{equation}
    The ASR objective $\mathcal{L}_{ASR}$ is formulated to maximize sequence-level log-likelihood, and $\mathcal{L}_{ASR}(\theta_{SLU}) = \mathcal{L}_{ASR}(\theta_{ASR})$.
    Before writing down $\mathcal{L}_{NLU}$, we describe a masking operation because ASR and BERT typically employ different subword tokenization methods. 
    
    \vspace{1mm}\noindent
    \textbf{Differentiate Through Subword Tokenizations}
    To concatenate $\theta_{ASR}$ and $\theta_{BERT}$ outputs along the hidden dimension, we need to make sure they have the same length along the token dimension. 
    We stored the first indices where $\bm{W}$ are broken down into subword tokens into a matrix: $M^{a}\in{\rm I\!R}^{N^{a}\times N}$ for $\theta_{ASR}$ and $M^{b}\in{\rm I\!R}^{N^{b}\times N}$ for $\theta_{BERT}$, 
    where $N$ is the number of tokens for $\bm{W}$ and $\bm{S}$, $N^{a}$ is the number of ASR subword tokens, and $N^{b}$ for BERT. 
    Let $H^{a}$ be the $\theta_{ASR}$ output matrix before softmax, and similarly $H^{b}$ for $\theta_{BERT}$. 
    The concatenated matrix $H^{cat}\in{\rm I\!R}^{N\times(F_{a}+F_{b})}$ is given as $H^{cat} = \text{concat}(\left[ (M^{a})^T H^{a}, (M^{b})^T H^{b} \right], \text{dim=1})$, 
    where $F_{a}$ and $F_{b}$ are hidden dimensions for $\theta_{ASR}$ and $\theta_{BERT}$.
    $\mathcal{L}_{NLU}$ is then,  
        \begin{align}
        \vspace{-1.5mm}
        \label{eq:e2E_nlu}
        \mathcal{L}_{NLU} &= 
        \mathop{\mathbb{E}_{}}
        \Big[
        \ln P(\bm{S}\mid H^{cat};\theta_{SL}) + 
        \ln P(\bm{I}\mid H^{cat};\theta_{IC})
        \Big],
        \vspace{-1.5mm}
        \end{align}
    where the sum of cross entropy losses for IC and SL are maximized, and $\theta_{ASR}$ and $\theta_{BERT}$ are updated through $H^{cat}$.
    Ground truth $\bm{W}$ is used as input to $\theta_{BERT}$ instead of $\bm{W}^*$ due to teacher forcing.    
        
\vspace{-2mm}
\subsection{Inference}
\label{subsec:inference}
\vspace{-1.5mm}
At test time, an input audio sequence $\bm{a} = a_{1:T}$ and the sets of all possible word tokens $\mathcal{W}$, slots $\mathcal{S}$, and intents $\mathcal{I}$ are given.
We are then interested in decoding for its target word sequence $\bm{w}^* = w_{1:N}$, slots sequence $\bm{s}^* = s_{1:N}$, and intent label $\bm{i}^*$.
Having obtained $\theta_{SLU}^*$, the decoding procedure for the end-to-end approach is,
        \begin{equation}
        \vspace{-2mm}
        \bm{w}^* = 
        \underset{w_{n}\in\mathcal{W}}{\text{argmax}}\prod_{n=1}^N p(w_{n}\mid w_{n-1}, \bm{a}; \theta_{ASR}^*), 
        \label{eq:e2e_inf_w}
        \end{equation}
        \vspace{-2mm}
        \begin{equation}
        \bm{i}^*,\bm{s}^* = 
        \underset{i\in\mathcal{I}}{\text{argmax }} p(i\mid w^*, \bm{a}; \theta_{SLU}^*),
        \underset{s_{n}\in\mathcal{S}}{\text{argmax}}\prod_{n=1}^N p(s_{n}\mid w^*, \bm{a}; \theta_{SLU}^*)
        \label{eq:e2e_inf_s}
        \end{equation}
This two step decoding procedure, first $\bm{w}^*$ then $(\bm{i}^*,\bm{s}^*)$ is necessary given that no explicit serialization on $\bm{W}$ and $\bm{S}$ are imposed, as in \cite{haghani2018audio,tomashenko2019recent}. 
While decoding for $(\bm{i}^*,\bm{s}^*)$, additional input $\bm{a}$ is given and we have $\bm{w}^*$ instead of $w_{n}$ given the context from self-attention in BERT. 
Note that here and throughout the work, we only take the top-1 hypothesis $\bm{w}^*$ (instead of top-N) to decode for $(\bm{i}^*,\bm{s}^*)$.

\section{Learning with Less Supervision}
\label{sec:wav2vec}
\vspace{-2mm}

\begin{figure}[!htbp]
    \centering
    \includegraphics[width=0.7\linewidth]{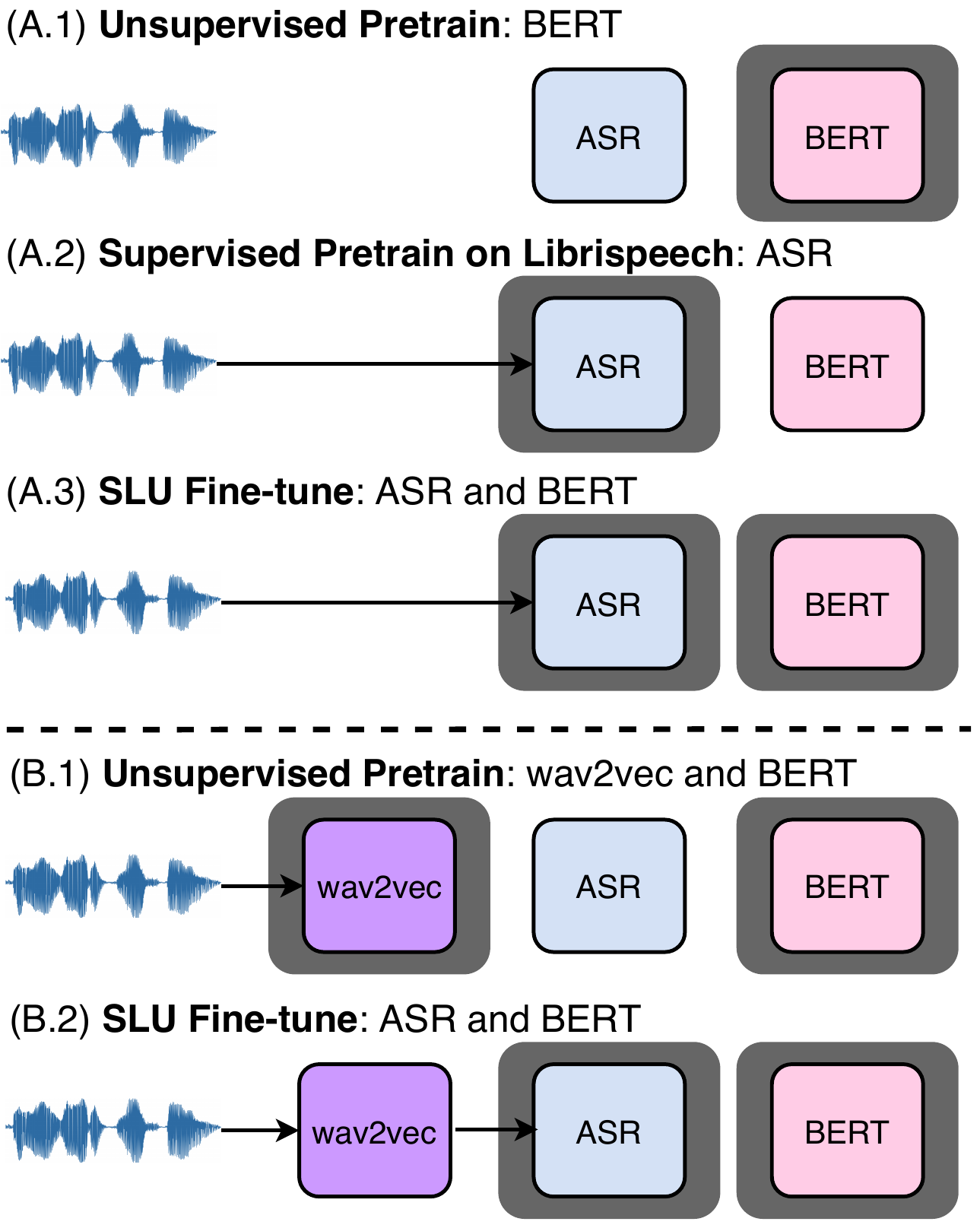}
    \vspace{-3mm}
    \caption{Two semi-supervised settings:
    \textbf{(A)} additional transcribed speech is available. 
    $\theta_{ASR}$ is pretrained and fine-tuned for ASR.
    \textbf{(B)} additional audio is available but without transcription.
    $\theta_{ASR}$ encoder is replaced with a pretrained wav2vec \cite{schneider2019wav2vec,baevski2020wav2vec} before fine-tuning.}
    \label{fig:wav2vec_learning}
    \vspace{-4mm}
    \end{figure}

    Our semi-supervised framework relies on pretrained ASR and NLU components. 
    Depending on the accessibility of the data, we explored two level of supervision\footnote{In either settings, the amount of IC/SL annotations remains the same.}. 
    The first setting is where an external transcribed corpus is available, and we utilized transfer learning for initializing the ASR.
    The second setting is where external audio is available but not transcriptions, and in this case, the ASR is initialized with self-supervised learning. 
    In both settings, BERT is pretrained with MLM and NSP as described in \cite{devlin2018bert}.
    Figure~\ref{fig:wav2vec_learning} distinguishes the two learning settings.
    
    \vspace{-3mm}

    \vspace{-3mm}
    \subsection{Transfer Learning from a Pretrained ASR}
    \label{subsec:transfer_learning}
    \vspace{-2mm}
    Following \cite{lugosch2019speech,rao2020speech,tomashenko2019recent}, $\theta_{ASR}$ is pretrained on an external transcribed speech corpus before fine-tuning on the target SLU dataset. 
    
    \vspace{-3mm}
    \subsection{Unsupervised ASR Pretraining with wav2vec}
    \label{subsec:wav2vec}
    \vspace{-2mm}
    According to UNESCO, 43\% of the languages in the world are endangered. 
    Supervised pretraining is not possible for many languages, as transcribing a language requires expert knowledge in phonetics, morphology, syntax, and so on.
    This partially motivates the line of self-supervised learning work in speech, that powerful learning representations require little fine-tuning data. 
    Returning to our topic, we asked, how does self-supervised learning help with learning semantics? 
    
    Among many others, wav2vec 1.0 \cite{schneider2019wav2vec} and 2.0 \cite{baevski2020wav2vec} demonstrated the effectiveness of self-supervised representations for ASR. 
    They are pretrained with contrastive losses \cite{oord2018representation}, and differ mainly by their architectures.  
    We replaced $\theta_{ASR}$ encoder with these wav2vec features, and appended the $\theta_{ASR}$ decoder for fine-tuning on SLU.

\vspace{-2.5mm}
\section{Experiments}
\label{sec:exp}

    \vspace{-2mm}
    \noindent
    \textbf{Datasets}
    ATIS \cite{hemphill1990atis} contains 8hr of audio recordings of people making flight reservations with corresponding human transcripts. 
    A total of 5.2k utterances with more than 600 speakers are present. 
    Note that ATIS is considerably smaller than those in-house SLU data used in \cite{coucke2018snips,haghani2018audio,serdyuk2018towards,rao2020speech}, justifying our limited semantics labels setup.
    Waveforms are sampled at 16kHz.
    For the unlabeled semantics data, we selected Librispeech 960hr (LS-960) \cite{panayotov2015librispeech} for pretraining. 
    Besides the original ATIS, models are evaluated on its noisy copy (augmented with MS-SNSD \cite{reddy2019scalable}).
    We made sure the noisy train and test splits in MS-SNSD do not overlap. 
    Text normalization is applied on the ATIS transcription with an open-source software\footnote{\url{https://github.com/EFord36/normalise}}. 
    Utterances are ignored if they contain words with multiple possible slot tags.

    \noindent
    \textbf{Hyperparameters}
    All speech is represented as sequences of 83-dimensional Mel-scale filter bank energies with $F_0$, computed every 10ms.
    Global mean normalization is applied. 
    E2E ASR is implemented in ESPnet, where it has 12 Transformer encoder layers and 6 decoder layers.
    The choice of the Transformer is similar to \cite{radfar2020end}.
    E2E ASR is optimized with hybrid CTC/attention losses \cite{watanabe2017hybrid} with label smoothing.
    The decoding beam size is set to 5 throughout this work. 
    We \textbf{do not} use an external LM during decoding.
    SpecAugment \cite{park2019specaugment} is used as the default for data augmentation.  
    A SentencePiece (BPE) vocabulary size is set to 1k.
    BERT is a bert-base-uncased from HuggingFace. 
    Code will be made available\footnote{\href{https://github.com/jefflai108/Semi-Supervsied-Spoken-Language-Understanding-PyTorch}{Code: Semi-Supervsied-Spoken-Language-Understanding-PyTorch}}.
    
    \vspace{-3.5mm}
    \subsection{E2E Evaluation with Slots Edit $F_{1}$ score.}
    \vspace{-2mm}
    Our framework is evaluated with an end-to-end evaluation metric, termed the slots edit $F_{1}$.
    Unlike slots $F_{1}$ score, slots edit $F_{1}$ accounts for instances where predicted sequences have different lengths as the ground truth.  
    It bears similarity with the E2E metric proposed in \cite{haghani2018audio,rao2020speech}.
    To calculate the score, the predicted text and oracle text are first aligned.
    For each slot label $v\in\mathcal{V}$, where $\mathcal{V}$ is the set of all possible slot labels except for the ``O" tag, we calculate the \underline{insertion} (false positive, FP), \underline{deletion} (false negative, FN) and \underline{substition} (FN and FP) of its slots value. 
    Slots edit $F_{1}$ is the harmonic mean of precision and recall over all slots:
    \vspace{0mm}
    \begin{equation}
    \label{eq:slot_edit_f1}
    \text{slots edit } F_{1} = \frac{\sum_{v\in\mathcal{V}} 2\times \text{TP}_{v}}
    {
    \sum_{v\in\mathcal{V}}\Big[ (2\times \text{TP}_{v}) + \text{FP}_{v} + \text{FN}_{v}\Big]
    }
    \end{equation} 

    \vspace{-5mm}
    \subsection{End-to-End 2-Stage Fine-tuning}
    \label{subsec:e2e_fine_tune}
    \vspace{-2mm}
    An observation from the experiment was that ASR is much harder than IC/SL. 
    Therefore, we adjusted our end-to-end training to a two-stage fine-tuning: 
    pretrain ASR on LS-960, then fine-tine ASR on ATIS, and lastly jointly fine-tune for ASR and IC/SL on ATIS.
    
        \vspace{-5mm}
        \begin{figure}[!]
        \centering
        \caption{Our SpeechBERT \cite{chuang2019speechbert} pretraining and fine-tuning setup.}
        \includegraphics[width=1.\linewidth]{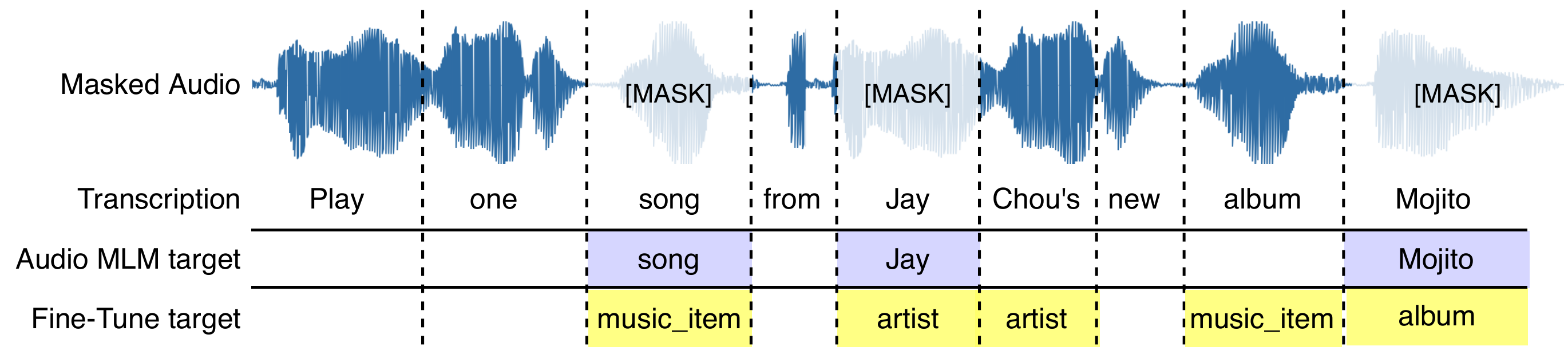}
        \label{fig:audio_mlm}
        \vspace{-8mm}
        \end{figure}
    
    \vspace{1.5mm} 
    \subsection{Baselines: Alternative $\theta_{SLU}$ Formulations}
    \label{subsec:baselines}
    \vspace{-1.5mm}
    Two variations for constructing $\theta_{SLU}$ are presented (refer to Figure~\ref{fig:overview}).
    They will be the baselines to the end-to-end approach. 
    
    \vspace{1mm}\noindent
    \textbf{2-Stage: Cascade ASR to BERT}
    A natural complement to the E2E approach is to \textit{separately} pretrain and fine-tune ASR and BERT. In this case, errors from $\bm{S}$ and $\bm{I}$ cannot be back-propagated to $\theta_{ASR}$.
    
    \vspace{1mm}\noindent
    \textbf{SpeechBERT : BERT in Speech-Text Embed Space}
    Another sensible way to construct $\theta_{SLU}$ is to somehow ``adapt" BERT such that it can take audio as input and outputs IC/SL, while not compromising its original semantic learning capacity. 
    SpeechBERT \cite{chuang2019speechbert} was initially proposed for spoken question answering, but we found the core idea of training BERT with audio-text pairs fitting as another baseline for our end-to-end approach. 
    We modified the pretraining and fine-tuning setup described in \cite{chuang2019speechbert} for SLU. 
    Audio MLM (c.f MLM in BERT \cite{devlin2018bert}) pretrains $\theta_{BERT}$ by mapping \textit{masked audio segments} to text.
    This pretraining step gradually adapts the original BERT to a phonetic-semantic joint embedding space. 
    Then, $\theta_{NLU}$ is fine-tuned by mapping \textit{unmasked audio segments} to IC/SL. Figure~\ref{fig:audio_mlm} illustrates the audio-text and audio-IC/SL pairs for SpeechBERT. 
    Unlike the end-to-end approach, $\theta_{ASR}$ is kept frozen throughout SpeechBERT pretraining and fine-tuning.
    
    \vspace{-4mm}
    \subsection{Main Results on Clean ATIS}
    \vspace{-2mm}
    We benchmarked our proposed framework with several prior works, and Table~\ref{tab:main_results} presents their WER, slots edit F1 and intent F1 results. 
    JointBERT \cite{chen2019bert} is our NLU baseline, where BERT is jointly fine-tuned for IC/SL, and it gets around 95\% slots edit $F_{1}$ and over 98\% IC F1. 
    Since JointBERT has access to the oracle text, this is the upper bound for our SLU models with speech as input. 
    CLM-BERT \cite{caostyle} explored using in-house conversational LM for NLU.
    We replicated \cite{tomashenko2019recent}, where an LAS \cite{chan2016listen} directly predicts interleaving word and slots tokens (serialized output), and optimized with CTC over words and slots.
    We also experimented with a Kaldi hybrid ASR.
    
    Both our proposed end-to-end and baselines approaches surpassed prior SLU work. 
    We hypothesize the performance gain originates from our choices of (1) adopting pretrained E2E ASR and BERT, (2) applying text-norm on target transcriptions for training the ASR, and (3) end-to-end fine-tuning text and IC/SL. 
    
    \vspace{-2mm}
    \begin{table}[!htbp]
    \vspace{-3mm}
    \caption{WER, slots edit and intent $F_{1}$ on ATIS. 
    ASR is pretrained on Librispeech 960h (LS-960).
    Results indicate our semi-supervised framework is effective in data scarcity setting, exceeding prior work in WER and IC/SL while approaching the NLU upperbound.}
    \vspace{2mm}
    \label{tab:main_results}
    \centering 
    \resizebox{\linewidth}{!}{
    \begin{tabular}{lcccc}
    \toprule
    \multirow{2}{*}{Frameworks} & Unlabeled & \multicolumn{3}{c}{ATIS clean test}\\
    \cmidrule(lr){3-5}
    {} & {Semantics Data} & {WER} & {slots edit $F_{1}$} & {intent $F_{1}$}\\
    \midrule
    \midrule
    \multicolumn{5}{l}{\textbf{NLU with Oracle Text}}\\
    JointBERT \cite{chen2019bert}       &           & -     & 95.64         & 98.99 \\
    
    \midrule
    \multicolumn{5}{l}{\textbf{Proposed}}\\
    End-to-End w/ 2-Stage               & LS-960    & 2.18  & \bf{95.88}    & 97.26\\
    2-Stage Baseline                    & LS-960    & 1.38  & 93.69         & 97.01\\
    SpeechBERT Baseline                 & LS-960    & 1.4   & 92.36         & \bf{97.4}\\
    
    \midrule 
    \multicolumn{5}{l}{\textbf{Prior Work}}\\
    ASR-Robust Embed \cite{huang2020learning}   & WSJ       & 15.55 & -         & 95.65 \\
    Kaldi Hybrid ASR+BERT                       & LS-960    & 13.31 & 85.13     & 94.56 \\
    ASR+CLM-BERT \cite{caostyle}                    & in-house  & 18.4. & 93.8\footnotemark  & 97.1\\
    LAS+CTC \cite{tomashenko2019recent}         & LS-460    & 8.32  & 86.85     & - \\
    
    \bottomrule
    \end{tabular}}
    \vspace{-3mm}
    \end{table}
    \addtocounter{footnote}{0}
    \footnotetext{For \cite{caostyle}, model predictions are evaluated only if its ASR hypothesis and human transcription have the same number of tokens. }
    % \footnote{shit shit shit }

    \vspace{-4mm}
    \subsection{Environmental Noise Augmentation}
    \label{subsec:nosie_fine_tune}
    \vspace{-2mm}
    A common scenario where users utter their spoken commands to SLU engines is when environmental noises are present in the background. 
    Nonetheless, common SLU benchmarking datasets like ATIS, SNIPS \cite{coucke2018snips}, or FSC \cite{lugosch2019speech} are very clean.
    To quantify model robustness under noisy settings, we augmented ATIS with environmental noise from MS-SNSD.
    Table~\ref{tab:noise_results} reveals that those work well on clean ATIS may break under realistic noises, and although our models are trained with SpecAugment, there is still a 4-27\% performance drop from clean test.  
    
    We followed the noise augmentation protocol in \cite{reddy2019scalable}, where for each sample, five noise files are sampled and added to the clean file with SNR levels of $[0, 10, 20, 30, 40]$dB, resulting in a five-fold augmentation.
    We observe that augmenting the training data with a diverse set of environmental noises work well, and there is now minimal model degradation.
    Our end-to-end approach reaches 95.46\% for SL and 97.4\% for IC, which is merely a 1-2\% drop from clean test, and almost a 40\% improvement over hybrid ASR+BERT. 

    \vspace{-5.5mm}
    \begin{table}[!htbp]
    \caption{Noise augmentation effectively reduces model degradation.}
    \vspace{.5mm}
    \label{tab:noise_results}
    \centering 
    \resizebox{0.8\linewidth}{!}{
    \begin{tabular}{lccc}
    \toprule
    \multirow{2}{*}{Frameworks} & \multicolumn{3}{c}{ATIS noisy test} \\
    \cmidrule(lr){2-4}
    {} & {WER} & {slots edit $F_{1}$} & {intent $F_{1}$}\\
    \midrule
    \midrule
    Kaldi Hybrid ASR+BERT                & 44.72 & 69.55 & 88.94 \\    
        
    \midrule
    \multicolumn{4}{l}{\textbf{Proposed w/ Noise Aug.}}\\
    End-to-End w/ 2-Stage           & 3.6 & \bf{95.46} & \bf{97.40} \\
    2-Stage Baseline                & 3.5 & 92.52 & 96.49 \\
    SpeechBERT Baseline             & 3.6 & 88.7 & 96.15 \\

    \midrule
    \multicolumn{4}{l}{\textbf{Proposed w/o Noise Aug.}}\\
    End-to-End w/ 2-Stage           & 9.62 & 91.54 & 96.14 \\
    2-Stage Baseline                & 8.98 & 90.09 & 95.74 \\
    SpeechBERT Baseline             & 9.0 & 81.72 & 94.05 \\

    \bottomrule
    \end{tabular}}
    \vspace{-6mm}
    \end{table}

\vspace{-1mm}
\subsection{Effectiveness of Unsupervised Pretraining with wav2vec}
\label{subsec:effect_wav2vec}
    \vspace{-2mm}
    Table~\ref{tab:wav2vec_results} shows the results on different ASR pretraining strategies: unsupervised pretraining with wav2vec, transfer learning from ASR, and no pretraining at all.
    We extracted both the latent vector $\bm{z}$ and context vector $\bm{c}$ from wav2vec 1.0.
    To simplify the pipeline and in contrast to \cite{schneider2019wav2vec}, we pre-extracted the wav2vec features and \textbf{did not} fine-tune wav2vec with $\theta_{SLU}$ on ATIS.
    We also chose not to decode with a LM to be consistent with prior SLU work. 
    We first observed the high WER for latent vector $\bm{z}$ from wav2vec 1.0, indicating they are sub-optimal and merely better than training from scratch by a slight margin.
    Nonetheless, encouragingly, context vector $\bm{c}$ from wav2vec 1.0 gets 67\% slots and 90\% intent $F_{1}$. 
    
    To improve the results, we added subsampling layers \cite{chan2016listen} on top of the wav2vec features to downsample the sequence length with convolution.
    The motivation here is $\bm{c}$ and $\bm{z}$ are comparably longer than the normal ASR encoder outputs.
    With sub-sampling, $\bm{c}$ from wav2vec 1.0 now achieves 85.64\% for SL and 95.67\% for IC, a huge relative improvement over training ASR from scratch, and closes the gap between unsupervised and supervised pretraining for SLU. 
    
    \vspace{-5mm}
    \begin{table}[!htbp]
    \caption{Effectiveness of different ASR pretraining strategies for our 2-Stage baseline.
    Results with wav2vec 2.0 \cite{baevski2020wav2vec} is omitted since they are not much better. 
    Setup is visaulized in Figure~\ref{fig:wav2vec_learning}.}
    \vspace{2mm}
    \label{tab:wav2vec_results}
    \centering 
    \resizebox{1.\linewidth}{!}{
    \begin{tabular}{lccc}
    \toprule
    \multirow{2}{*}{Frameworks} & \multicolumn{3}{c}{ATIS clean test} \\
    \cmidrule(lr){2-4}
    {} & {WER} & {slots edit $F_{1}$} & {intent $F_{1}$}\\
    \midrule
    \midrule

    \multicolumn{4}{l}{\textbf{Proposed 2-Stage w/o ASR Pretraining}}\\
    2-Stage Baseline                        & 58.7  & 29.22     & 82.08\\
    \midrule
    \multicolumn{4}{l}{\textbf{Proposed 2-Stage w/ Transfer Learning from ASR}}\\
    2-Stage Baseline                        & 1.38  & 93.69     & 97.01\\
    \midrule
    \multicolumn{4}{l}{\textbf{Proposed 2-Stage w/ Unsupervised Pretraining}}\\
    wav2vec1.0 \cite{schneider2019wav2vec} $\bm{z}$ + 2-Stage               & 54.2  & 35.04     & 83.68\\
    wav2vec1.0 \cite{schneider2019wav2vec} $\bm{c}$ + 2-Stage               & 30.4  & 67.33     & 89.86\\
    wav2vec1.0 \cite{schneider2019wav2vec} $\bm{c}$ + subsample + 2-Stage   & 13.2  & \bf{85.64}    & \bf{95.67}\\
    % wav2vec2.0 \cite{baevski2020wav2vec} $\bm{c}$ + 2-Stage                 & 76.8  & 11.59     & 74.54\\
    % wav2vec2.0 \cite{baevski2020wav2vec} $\bm{c}$ + subsample +  2-Stage    & 44.1  & 47.72     & 84.92\\
    
    \bottomrule
    \end{tabular}}
    \vspace{-7mm}
    \end{table}

\section{Conclusions and Future Work}
\label{sec:conclusion}
\vspace{-2mm}
This work attempts to respond to a classic paper "What is left to be understood in ATIS? \cite{tur2010left}", and to the advancement put forward by contextualized LM and end-to-end SLU up against semantics understanding. 
We showed for the first time that an SLU model with speech as input could perform on par with NLU models on ATIS, entering the 5\% ``corpus errors" range \cite{tur2010left,bechet2018atis}. 
However, we collectively believe that there are unsolved questions remaining, such as the prospect of building a single framework for \textbf{multi-lingual SLU} \cite{glass1995multilingual}, or the need for a more spontaneous SLU corpus that is not limited to short segments of spoken commands. 

\noindent
{\footnotesize
\textbf{Acknowledgments}
\label{sec:acknowledgment}
We thank Nanxin Chen, Erica Cooper, Alexander H. Liu, Wei Fang, and Fan-Keng Sun for their comments on this work.}

\vfill\pagebreak

{\footnotesize
\bibliographystyle{IEEEbib}
\bibliography{main}}

\end{document}